\documentclass[10pt,onecolumn]{article}
\usepackage{times}
\usepackage[dvips]{graphicx}
\usepackage[hang,scriptsize]{subfigure}
\usepackage{latexsym}

\usepackage{amsmath} 
\usepackage{amssymb} 
\usepackage{graphics} 
\usepackage{listings}    

\makeatletter
\def\@maketitle{%
  \vbox to 9cm{
    \hsize\textwidth
    \linewidth\hsize
    \vspace{1.5cm}
    \centering
    {\bfseries\LARGE \@title \par}
    \vspace{12pt}
    {\fontsize{11pt}{13pt}\selectfont \begin{tabular}[t]{c}\@author \end{tabular}\par}
    \vfill} 
}
\renewcommand\section{\@startsection{section}{1}{\z@}%
                       {-12\p@ \@plus -4\p@ \@minus -4\p@}%
                       {6\p@ \@plus 4\p@ \@minus 4\p@}%
                       {\normalfont\large\bfseries
                        \rightskip=\z@ \@plus 8em\pretolerance=10000 }}
\renewcommand\subsection{\@startsection{subsection}{2}{\z@}%
                       {-12\p@ \@plus -4\p@ \@minus -4\p@}%
                       {6\p@ \@plus 4\p@ \@minus 4\p@}%
                       {\normalfont\fontsize{11pt}{13pt}\selectfont\bfseries
                        \rightskip=\z@ \@plus 8em\pretolerance=10000 }}
\renewcommand\subsubsection{\@startsection{subsubsection}{3}{\z@}%
                       {-12\p@ \@plus -4\p@ \@minus -4\p@}%
                       {6\p@ \@plus 4\p@ \@minus 4\p@}%
                       {\normalfont\normalsize\itshape}}
\renewcommand\paragraph{\@startsection{paragraph}{4}{\z@}%
                       {-12\p@ \@plus -4\p@ \@minus -4\p@}%
                       {-0.5em \@plus -0.22em \@minus -0.1em}%
                       {\normalfont\normalsize\itshape}}
\makeatother

\renewenvironment{abstract}%
  {\small
    \list{}{\labelwidth0pt
      \leftmargin0pt \rightmargin\leftmargin
      \listparindent\parindent \itemindent0pt
      \parsep0pt
      }%
    \item[\hskip\labelsep\bfseries\abstractname\enspace --] \itshape}{\endlist}

\newcommand{\keywordsname}{Keywords}
\newenvironment{keywords}%
  {\small
    \list{}{\labelwidth0pt
      \leftmargin0pt \rightmargin\leftmargin
      \listparindent\parindent \itemindent0pt
      \parsep0pt
      }%
    \item[\hskip\labelsep\bfseries\keywordsname:]}{\endlist}

\begin{document}

\pagestyle{myheadings}
\markboth{Proceedings of Fusion 2004 International  Conference,  Stockholm, Sweden, June 28-July1, 2004, - \copyright ISIF}{Proceedings of Fusion 2004 International  Conference,  Stockholm, Sweden, June 28-July1, 2004, - \copyright ISIF}

\title{The Combination of Paradoxical, Uncertain and Imprecise 
Sources of Information based on DSmT\\
and Neutro-Fuzzy Inference}

\author{\begin{tabular}{c@{\extracolsep{4em}}c@{\extracolsep{1em}}c}
{\bf Florentin Smarandache} & {\bf Jean Dezert}\\
Dept. of Mathematics &  ONERA/DTIM/IED \\
Univ. of New Mexico  & 29 Av. de la  Division Leclerc \\
Gallup, NM 8730 & 92320 Ch\^{a}tillon \\
U.S.A.  & France \\
{\tt smarand@unm.edu} & {\tt Jean.Dezert@onera.fr}
\end{tabular}}
\vspace{-2cm}

\date{}
\maketitle
\thispagestyle{empty}
\pagestyle{empty}

\begin{abstract}
The management and combination of uncertain, imprecise, fuzzy and even paradoxical or high conflicting sources of information has always been, and still remains today, of primal importance for the development of reliable modern information systems involving artificial reasoning. In this chapter, we present a survey of our recent theory of plausible and paradoxical reasoning, known as Dezert-Smarandache Theory (DSmT) in the literature, developed for dealing with imprecise, uncertain and paradoxical sources of information. We focus our presentation here rather on the foundations of DSmT, and on the two important new rules of combination, than on browsing specific applications of DSmT available in literature. Several simple examples are given throughout the presentation to show the efficiency and the generality of this new approach. The last part of this chapter concerns the presentation of the neutrosophic logic, the neutro-fuzzy inference and its connection with DSmT.  Fuzzy logic and neutrosophic logic are useful tools in decision making after fusioning the information using the DSm hybrid rule of combination of masses.
\end{abstract}


\section{Introduction}

The management and combination of uncertain, imprecise, fuzzy and even paradoxical or high conflicting sources of information has always been, and still remains today, of primal importance for the development of reliable modern information systems involving artificial reasoning. The combination (fusion) of information arises in many fields of applications nowadays (especially in defense, medicine, finance, geo-science, economy, etc). When several sensors, observers or experts have to be combined together to solve a problem, or if one wants to update our current estimation of solutions for a given problem with some new information available, we need powerful and solid mathematical tools for the fusion, specially when the information one has to deal with is imprecise and uncertain. In this chapter, we present a survey of our recent theory of plausible and paradoxical reasoning, known as Dezert-Smarandache Theory (DSmT) in the literature, developed for dealing with imprecise, uncertain and paradoxical sources of information. Recent publications have shown the interest and the ability of DSmT to solve problems where other approaches fail, especially when conflict between sources becomes high. We focus our presentation here rather on the foundations of DSmT, and on the two important new rules of combination, than on browsing specific applications of DSmT available in literature. A particular attention is given to general (hybrid) rule of combination which deals with any model for fusion problems, depending on the nature of elements or hypotheses involved into them. The Shafer's model on which is based the Dempster-Shafer Theory (DST) appears only as a specific DSm hybrid model and can be easily handled by our approach as well. Several simple examples are given throughout the presentation to show the efficiency and the generality of this new approach. The last part of this work concerns the presentation of the neutrosophic logic, the neutro-fuzzy inference and its connection with DSmT.  Fuzzy logic and neutrosophic logic are useful tools in decision making after fusioning the information using the DSm hybrid rule of combination of masses.

\section{Foundations of the  DSmT}

The development of the DSmT (Dezert-Smarandache Theory of plausible and paradoxical reasoning \cite{DSmTBook_2004a}) arises from the necessity to overcome the inherent limitations of the DST (Dempster-Shafer Theory \cite{Shafer_1976}) which are closely related with the acceptance of Shafer's model for the fusion problem under consideration (i.e. the frame of {\it{discernment}} $\Theta$ defined as a finite set of {\it{exhaustive}} and {\it{exclusive}} hypotheses $\theta_i$, $i=1,\ldots,n$), the third middle excluded principle (i.e. the existence of the complement for any elements/propositions belonging to the power set of $\Theta$), and the acceptance of Dempter's rule of combination (involving normalization) as the framework for the combination of independent sources of evidence. Discussions on limitations of DST and presentation of some alternative rules to the Dempster's rule of combination can be found in \cite{Zadeh_1979,Zadeh_1984,Zadeh_1985,Yager_1985,Zadeh_1986,Dubois_1986c,Yager_1987,Pearl_1988,Smets_1988,Voorbraak_1991,Inagaki_1991,Murphy_2000,Lefevre_2002,Sentz_2002,Lefevre_2003,DSmTBook_2004a} and therefore they will be not reported in details in this chapter due to space limitation. We argue that these three fundamental conditions of the DST can be removed and another new mathematical approach for combination of evidence is possible.\\

The basis of the DSmT is  the refutation of the principle of the third excluded middle and Shafer's model, since for a wide class of fusion problems the intrinsic nature of hypotheses can be only vague and imprecise in such a way that precise refinement is just impossible to obtain in reality so that the exclusive elements $\theta_i$ cannot be properly identified and precisely separated. Many problems involving fuzzy continuous and relative concepts described in natural language and having no absolute interpretation like tallness/smallness, pleasure/pain, cold/hot, Sorites paradoxes, etc,  enter in this category. DSmT starts with the notion of {\it{free DSm model}}, denoted $\mathcal{M}^f(\Theta)$, and considers $\Theta$ only as a frame of exhaustive elements $\theta_i$, $i=1,\ldots,n$ which can potentially overlap. This model is {\it{free}} because no other assumption is done on the hypotheses, but the weak exhaustivity constraint which can always been satisfied according the closure principle explained in \cite{DSmTBook_2004a}. No other constraint is involved in the free DSm model.
When the free DSm model holds, the classic commutative and associative DSm rule of combination (corresponding to the conjunctive consensus defined on the free Dedekind's lattice) is performed.\\

Depending on the intrinsic nature of the elements of the fusion problem under consideration, it can however happen that the free model does not fit the reality because some subsets of $\Theta$ can contain elements known to be truly exclusive but also truly non existing at all at a given time (specially when working on dynamic fusion problem where the frame $\Theta$ varies with time with the revision of the knowledge available). These integrity constraints are then explicitly and formally introduced into the free DSm model $\mathcal{M}^f(\Theta)$ in order to adapt it properly to fit as close as possible with the reality and permit to construct a {\it{hybrid DSm model}} $\mathcal{M}(\Theta)$ on which the combination will be efficiently performed. Shafer's model, denoted $\mathcal{M}^0(\Theta)$, corresponds to a very specific hybrid DSm model including all possible exclusivity constraints. The DST has been developed for working only with $\mathcal{M}^0(\Theta)$ while the DSmT has been developed for working with any kind of hybrid model (including Shafer's model and the free DSm model), to manage as efficiently and precisely as possible imprecise, uncertain and potentially high conflicting sources of evidence while keeping in mind the possible dynamicity of the information fusion problematic. The foundations of the DSmT are therefore totally different from those of all existing approaches managing uncertainties, imprecisions and conflicts. DSmT provides a new interesting way to attack the  information fusion problematic with a general framework in order to cover a wide variety of problems. \\

DSmT refutes also the idea that sources of evidence provide their beliefs with the same absolute interpretation of elements of the same frame $\Theta$ and the conflict between sources arises not only because of the possible unreliabilty of sources, but also because of possible different and relative interpretation of $\Theta$, e.g. what is considered as good for somebody can be considered as bad for somebody else. There is some unavoidable subjectivity in the belief assignments provided by the sources of evidence, otherwise it would mean that all bodies of evidence have a same objective and universal interpretation (or measure) of the phenomena under consideration, which unfortunately rarely occurs in reality, but when bba are based on some {\it{objective probabilities}} transformations. But in this last case, probability theory can handle properly and efficiently the information, and the DST, as well as the DSmT, becomes useless. If we now get out of the probabilistic background argumentation for the construction of bba, we claim that in most of cases, the sources of evidence provide their beliefs about elements of the frame of the fusion problem only based on their own limited knowledge and experience without reference to the (inaccessible) absolute truth of the space of possibilities. First applications of DSmT for target tracking, satellite surveillance, situation analysis and sensor allocation optimization can be found in \cite{DSmTBook_2004a}.

\subsection{Notion of hyper-power set $D^\Theta$}

One of the cornerstones of the DSmT is  the free Dedekind lattice \cite{Dedekind_1897} denoted {\it{hyper-power set}} in the DSmT framework. Let $\Theta=\{\theta_{1},\ldots,\theta_{n}\}$ be a finite set (called frame) of $n$ 
exhaustive elements\footnote{We do not assume here that elements $\theta_i$ are necessary exclusive. There is no restriction on $\theta_i$ but the exhaustivity.}. The hyper-power set $D^\Theta$ is defined as the set of all composite propositions built from elements of $\Theta$ with $\cup$ and $\cap$ operators\footnote{$\Theta$ generates $D^\Theta$ under operators $\cup$ and $\cap$} such that: 
\begin{enumerate}
\item $\emptyset, \theta_1,\ldots, \theta_n \in D^\Theta$.
\item  If $A, B \in D^\Theta$, then $A\cap B\in D^\Theta$ and $A\cup B\in D^\Theta$.
\item No other elements belong to $D^\Theta$, except those obtained by using rules 1 or 2.
\end{enumerate}
The dual (obtained by switching $\cup$ and $\cap$ in expressions) of $D^\Theta$ is itself.  There are elements in $D^\Theta$ which are self-dual (dual to themselves), for example $\alpha_8$ for the case when $n=3$ in the following example. The cardinality of $D^\Theta$ is majored by 
$2^{2^n}$ when the cardinality of $\Theta$ equals $n$, i.e. $\vert\Theta\vert=n$. The generation 
of hyper-power set $D^\Theta$ is closely related with the famous Dedekind problem \cite{Dedekind_1897,Comtet_1974} on enumerating the 
set of isotone Boolean functions. The generation of the hyper-power set is presented in \cite{DSmTBook_2004a}. Since for any given finite set $\Theta$, $\vert D^\Theta\vert  \geq \vert 2^\Theta\vert $ we call $D^\Theta$ the  {\it{hyper-power set}} of $\Theta$.\\

\noindent{\it{Example of the first hyper-power sets $D^\Theta$}}
\begin{itemize}
\item
For the degenerate case ($n=0)$ where $\Theta=\{  \}$, one has $D^\Theta=\{\alpha_0\triangleq\emptyset\}$ and $\vert D^\Theta\vert = 1$.
\item When $\Theta=\{\theta_{1}\}$, one has $D^\Theta=\{\alpha_0\triangleq\emptyset,\alpha_1\triangleq\theta_1 \}$ and $\vert D^\Theta\vert = 2$.
\item When $\Theta=\{\theta_{1},\theta_{2}\}$, one has
$D^\Theta=\{\alpha_0,\alpha_1,\ldots,\alpha_{4} \}$ and $\vert D^\Theta\vert = 5$ with
$\alpha_0\triangleq\emptyset$, $\alpha_1\triangleq\theta_1\cap\theta_2$, $\alpha_2\triangleq\theta_1$, $\alpha_3\triangleq\theta_2 $ and $\alpha_4\triangleq\theta_1\cup\theta_2 $.
\item
When $\Theta=\{\theta_{1},\theta_{2},\theta_{3}\}$,  one has $D^\Theta=\{\alpha_0,\alpha_1,\ldots,\alpha_{18} \}$ and $\vert D^\Theta\vert = 19$ with
\begin{equation*}
\begin{array}{ll}
\alpha_0\triangleq\emptyset          &                                       \\
\alpha_1\triangleq\theta_1\cap\theta_2\cap\theta_3    &\alpha_{10}\triangleq\theta_2   \\
\alpha_2\triangleq\theta_1\cap\theta_2    & \alpha_{11}\triangleq\theta_3                            \\
\alpha_3\triangleq\theta_1\cap\theta_3    &\alpha_{12}\triangleq(\theta_1\cap\theta_2)\cup\theta_3                            \\
\alpha_4\triangleq\theta_2\cap\theta_3    &\alpha_{13}\triangleq(\theta_1\cap\theta_3)\cup\theta_2                             \\
\alpha_5\triangleq(\theta_1\cup\theta_2)\cap\theta_3  & \alpha_{14}\triangleq(\theta_2\cap\theta_3)\cup\theta_1      \\
\alpha_6\triangleq(\theta_1\cup\theta_3)\cap\theta_2  & \alpha_{15}\triangleq\theta_1\cup\theta_2       \\
\alpha_7\triangleq(\theta_2\cup\theta_3)\cap\theta_1 & \alpha_{16}\triangleq\theta_1\cup\theta_3        \\
\alpha_8\triangleq(\theta_1\cap\theta_2)\cup(\theta_1\cap\theta_3)\cup(\theta_2\cap\theta_3)  & \alpha_{17}\triangleq\theta_2\cup\theta_3    \\
\alpha_9\triangleq\theta_1  & \alpha_{18}\triangleq\theta_1\cup\theta_2\cup\theta_3                                                   \end{array}
\end{equation*}
\end{itemize}

The cardinality of hyper-power set $D^\Theta$ for $n\geq1$ follows the sequence of Dedekind's numbers \cite{Sloane_2003}, i.e.
1,2,5,19,167, 7580,7828353,...  and analytical expression of Dedekind's numbers has been obtained recently by Tombak in \cite{Tombak_2001} (see \cite{DSmTBook_2004a} for details on generation and ordering of $D^\Theta$).

\subsection{Notion of free and hybrid DSm models}
\label{Sec:DSMmodels}

Elements $\theta_i$, $i=1,\ldots,n$ of $\Theta$ constitute the finite set of hypotheses/concepts characterizing the fusion problem under consideration. $D^\Theta$ constitutes what we call the {\it{free DSm model}} $\mathcal{M}^f(\Theta)$ and allows to work with fuzzy concepts which depict a continuous and relative intrinsic nature. Such kinds of concepts cannot be precisely refined in an absolute interpretation because of the unapproachable universal truth.\\

However for some particular fusion problems involving discrete concepts, elements $\theta_i$ are truly exclusive. In such case, all the exclusivity constraints on $\theta_i$, $i=1,\ldots,n$ have to be included in the previous model to characterize properly the true nature of the fusion problem and to fit it with the reality. By doing this, the hyper-power set $D^\Theta$ reduces naturally to the classical power set $2^\Theta$ and this constitutes the most restricted hybrid DSm model, denoted  $\mathcal{M}^0(\Theta)$, coinciding with Shafer's model. As an exemple, let's consider the 2D problem where $\Theta=\{\theta_1,\theta_2\}$ with $D^\Theta=\{\emptyset,\theta_1\cap\theta_2,\theta_1,\theta_2,\theta_1\cup\theta_2\}$ and assume now that $\theta_1$ and $\theta_2$ are truly exclusive (i.e. Shafer's model $\mathcal{M}^0$ holds), then because $\theta_1\cap\theta_2\overset{\mathcal{M}^0}{=}\emptyset$, one gets $D^\Theta=\{\emptyset,\theta_1\cap\theta_2\overset{\mathcal{M}^0}{=}\emptyset,\theta_1,\theta_2,\theta_1\cup\theta_2\}=\{\emptyset,\theta_1,\theta_2,\theta_1\cup\theta_2\}\equiv 2^\Theta$.\\

Between the class of fusion problems corresponding to the free DSm model $\mathcal{M}^f(\Theta)$ and  the class of fusion problems corresponding to Shafer's model $\mathcal{M}^0(\Theta)$, there exists another wide class of hybrid fusion problems involving in $\Theta$ both fuzzy continuous concepts and discrete hypotheses. In such (hybrid) class, some exclusivity constraints and possibly some non-existential constraints (especially when working on dynamic\footnote{i.e. when the frame $\Theta$ and/or the model $\mathcal{M}$ is changing with time.} fusion) have to be taken into account. Each hybrid fusion problem of this class will then be characterized by a proper hybrid DSm model $\mathcal{M}(\Theta)$ with $\mathcal{M}(\Theta)\neq\mathcal{M}^f(\Theta)$ and $\mathcal{M}(\Theta)\neq \mathcal{M}^0(\Theta)$. As simple example of DSm hybrid model, let's consider the 3D case with the frame $\Theta=\{\theta_1,\theta_2,\theta_3\}$ with the model $\mathcal{M}\neq\mathcal{M}^f$ in which we force all possible conjunctions to be empty, but $\theta_1\cap\theta_2$. This hybrid DSm model is then represented with  the following Venn diagram (where boundaries of intersection of $\theta_1$ and $\theta_2$ are not precisely defined if $\theta_1$ and $\theta_2$ represent only fuzzy concepts like {\it{smallness}} and {\it{tallness}} by example).
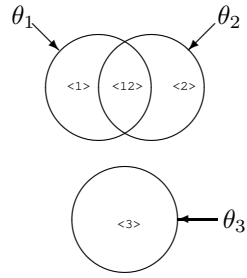
\begin{center}
{\tt \setlength{\unitlength}{1pt}
\begin{picture}(90,90)
\thinlines    
\put(40,60){\circle{40}}
\put(60,60){\circle{40}}
\put(50,10){\circle{40}}
\put(15,84){\vector(1,-1){10}}
\put(7,84){$\theta_{1}$}
\put(84,84){\vector(-1,-1){10}}
\put(85,84){$\theta_{2}$}
\put(85,10){\vector(-1,0){15}}
\put(87,7){$\theta_{3}$}
\put(45,59){\tiny{<12>}}
\put(47,7){\tiny{<3>}}
\put(68,59){\tiny{<2>}}
\put(28,59){\tiny{<1>}}
\end{picture}}
\end{center}

\vspace{1cm}

\subsection{Generalized belief functions}

From a general frame $\Theta$, we define a map $m(.): 
D^\Theta \rightarrow [0,1]$ associated to a given body of evidence $\mathcal{B}$ as 
\begin{equation}
m(\emptyset)=0 \qquad \text{and}\qquad \sum_{A\in D^\Theta} m(A) = 1 
\end{equation}
\noindent The quantity $m(A)$ is called the {\it{generalized basic belief assignment/mass}} (gbba) of $A$.\\

\noindent
The {\it{generalized belief and plausibility functions}} are defined in almost the same manner as within the DST, i.e.
\begin{equation}
\text{Bel}(A) = \sum_{\substack{B\subseteq A\\ B\in D^\Theta}} m(B)
\qquad\qquad\text{Pl}(A) = \sum_{\substack{B\cap A\neq\emptyset \\ B\in D^\Theta}} m(B)
\end{equation}

These definitions are compatible with the definitions of classical belief functions in the DST framework when $D^\Theta$ reduces to $2^\Theta$ for fusion problems where Shafer's model $\mathcal{M}^0(\Theta)$ holds. We still have $\forall A\in D^\Theta,\, \text{Bel}(A)\leq \text{Pl}(A)$.
Note that when working with the free DSm model $\mathcal{M}^f(\Theta)$, one has always $\text{Pl}(A) =1$ $\forall A\neq\emptyset \in D^\Theta$ which is normal.

\subsection{The classic DSm rule of combination}

When the free DSm model $\mathcal{M}^f(\Theta)$ holds for the fusion problem under consideration, the  classic DSm rule of combination $m_{\mathcal{M}^f(\Theta)}\equiv m(.)\triangleq [m_{1}\oplus m_{2}](.)$ of two independent\footnote{While independence is a difficult concept to define in all theories managing epistemic uncertainty, we follow here the interpretation of Smets in \cite{Smets_1986} and \cite{Smets_1988}, p. 285 and consider that two sources of evidence are independent (i.e distinct and noninteracting) if each leaves one totally ignorant about the particular value the other will take.} sources of evidences $\mathcal{B}_{1}$ and  $\mathcal{B}_{2}$ over the same 
frame $\Theta$ with belief functions $\text{Bel}_{1}(.)$ and 
 $\text{Bel}_{2}(.)$ associated with gbba $m_{1}(.)$ and $m_{2}(.)$ corresponds to the conjunctive consensus of the sources. It is  given by \cite{DSmTBook_2004a}:
 \begin{equation}
\forall C\in D^\Theta,\qquad m_{\mathcal{M}^f(\Theta)}(C) \equiv m(C) = 
 \sum_{\substack{A,B\in D^\Theta\\ A\cap B=C}}m_{1}(A)m_{2}(B)
 \label{JDZT}
 \end{equation}
 
Since $D^\Theta$ is closed under $\cup$ and $\cap$ set operators, this new rule 
of combination guarantees that $m(.)$ is a proper generalized belief assignment, i.e. $m(.): D^\Theta \rightarrow [0,1]$. This rule of combination is commutative and associative 
and can always be used for the fusion of sources involving fuzzy concepts when free DSm model holds for the problem under consideration. This rule can be directly and easily extended for the combination of $k > 2$ independent sources of evidence \cite{DSmTBook_2004a}.\\

This classic DSm rule of combination looks very expensive in terms of computations and memory size due to the huge number of elements in $D^\Theta$ when the cardinality of $\Theta$ increases. This remark is however valid only if the cores (the set of focal elements of gbba) $\mathcal{K}_1(m_1)$ and $\mathcal{K}_2(m_2)$ coincide with $D^\Theta$, i.e. when $m_1(A)>0$ and $m_2(A)>0$ for all $A\neq\emptyset\in D^\Theta$. Fortunately, it is important to note here that in most of the practical applications the sizes of $\mathcal{K}_1(m_1)$ and $\mathcal{K}_2(m_2)$ are much smaller than $\vert D^\Theta\vert$ because bodies of evidence generally allocate their basic belief assignments only over a subset of the hyper-power set. This makes things easier for the implementation of the classic DSm rule \eqref{JDZT}. The DSm rule is actually very easy to implement. It suffices for each focal element of $\mathcal{K}_1(m_1)$ to multiply it with the focal elements of $\mathcal{K}_2(m_2)$ and then to  pool all combinations which are equivalent under the  algebra of sets.\\

While very costly in term on merory storage in the worst case (i.e. when all $m(A)>0$, $A\in D^\Theta$ or $A\in2^{\Theta_{ref}}$), the DSm rule however requires much smaller memory storage than for the DST working on the ultimate refinement $2^{\Theta_{ref}}$ of same initial frame $\Theta$ as shown in following table
\begin{center}
\begin{tabular}{|l|l|l|}
\hline
$\vert\Theta\vert=n$ & $\vert D^\Theta \vert$ & $\vert 2^{\Theta_{ref}} \vert = 2^{2^n-1}$ \\
\hline
2 &  5 &  $2^3=8$ \\
3 &  19 &  $2^7=128$ \\
4 &  167 &  $2^{15}=32768$ \\
5 &  7580 &  $2^{31}=2147483648$\\
\hline
\end{tabular}
\end{center}
 
However in most fusion applications only a small subset of elements of $D^\Theta$ have a non null basic belief mass because all the commitments are just usually impossible to assess precisely when the dimension of the problem increases. Thus, it is not necessary to generate and keep in memory all elements of $D^\Theta$ or $2^{\Theta_{ref}}$ but only those which have a positive belief mass.  However there is a real technical challenge on how to manage efficiently all elements of the hyper-power set. This problem is obviously much more difficult when trying to work on the refined frame of discernment $2^{\Theta_{ref}}$ if one prefer to apply Dempster-Shafer theory and use the Dempster's rule of combination. It is important to keep in mind that the ultimate refined frame consisting in exhaustive and exclusive finite set of refined hypotheses is just impossible to justify and to define precisely for all problems dealing with fuzzy and ill-defined continuous concepts. A full discussion and example on refinement can be found in \cite{DSmTBook_2004a}.

\subsection{The hybrid DSm rule of combination}

When the free DSm model $\mathcal{M}^f(\Theta)$ does not hold due to the true nature of the fusion problem under consideration which requires to take into account some known integrity constraints, one has to work with a proper hybrid DSm model $\mathcal{M}(\Theta)\neq\mathcal{M}^f(\Theta)$. In such case, the hybrid DSm rule of combination based on the chosen hybrid DSm model $\mathcal{M}(\Theta)$ for $k\geq 2$ independent sources of information is defined for all $A\in D^\Theta$ as \cite{DSmTBook_2004a}:
\begin{equation}
m_{\mathcal{M}(\Theta)}(A)\triangleq 
\phi(A)\Bigl[ S_1(A) + S_2(A) + S_3(A)\Bigr]
 \label{eq:DSmHkBis1}
\end{equation}
\noindent
where $\phi(A)$ is the {\it{characteristic non-emptiness function}} of a set $A$, i.e. $\phi(A)= 1$ if  $A\notin \boldsymbol{\emptyset}$ and $\phi(A)= 0$ otherwise, where $\boldsymbol{\emptyset}\triangleq\{\boldsymbol{\emptyset}_{\mathcal{M}},\emptyset\}$. $\boldsymbol{\emptyset}_{\mathcal{M}}$ is the set  of all elements of $D^\Theta$ which have been forced to be empty through the constraints of the model $\mathcal{M}$ and $\emptyset$ is the classical/universal empty set. $S_1(A)\equiv m_{\mathcal{M}^f(\theta)}(A)$, $S_2(A)$, $S_3(A)$ are defined by 
\begin{equation}
S_1(A)\triangleq \sum_{\substack{X_1,X_2,\ldots,X_k\in D^\Theta\\ (X_1\cap X_2\cap\ldots\cap X_k)=A}} \prod_{i=1}^{k} m_i(X_i)
\end{equation}
\begin{equation}
S_2(A)\triangleq \sum_{\substack{X_1,X_2,\ldots,X_k\in\boldsymbol{\emptyset}\\ [\mathcal{U}=A]\vee [(\mathcal{U}\in\boldsymbol{\emptyset}) \wedge (A=I_t)]}} \prod_{i=1}^{k} m_i(X_i)
\end{equation}
\begin{equation}
S_3(A)\triangleq\sum_{\substack{X_1,X_2,\ldots,X_k\in D^\Theta \\ u(c(X_1\cap X_2\cap\ldots\cap X_k))=A \\ (X_1\cap X_2\cap \ldots\cap X_k)\in\boldsymbol{\emptyset}}}  \prod_{i=1}^{k} m_i(X_i)
\end{equation}
with $\mathcal{U}\triangleq u(X_1)\cup u(X_2)\cup \ldots \cup u(X_k)$ where $u(X)$ is the union of all $\theta_i$ that compose $X$, $I_t \triangleq \theta_1\cup \theta_2\cup\ldots\cup \theta_n$ is the total ignorance, and $c(X)$ is the canonical form\footnote{The canonical form is introduced here in order to improve the original formula given in \cite{DSmTBook_2004a} for preserving the neutral impact of the vacuous belief mass $m(\Theta)=1$ within complex hybrid models.} of $X$, i.e. its simplest form (for example if $X=(A\cap B)\cap (A\cup B\cup C)$, 
$c(X)=A\cap B$). $S_1(A)$ corresponds to the classic DSm rule for $k$ independent sources based on the free DSm model $\mathcal{M}^f(\Theta)$; $S_2(A)$ represents the mass of all relatively and absolutely empty sets which is transferred to the total or relative ignorances associated with non existential constraints (if any, like in some dynamic problems); $S_3(A)$ transfers the sum of relatively empty sets directly onto the canonical disjunctive form of non-empty sets.\\

The hybrid DSm rule of combination generalizes the classic DSm rule of combination and is not equivalent to Dempter's rule. It works for any models (the free DSm model, Shafer's model or any other hybrid models) when manipulating {\it{precise}} generalized (or eventually classical) basic belief functions. An extension of this rule for the combination of {\it{imprecise}} generalized (or eventually classical) basic belief functions is presented in next section.\\

Note that in DSmT framework it is also possible to deal directly with complements if necessary depending on the problem under consideration and the information provided by the sources of evidence themselves. The first and simplest way is to work on Shafer's model when utimate refinement is possible.
The second way is to deal with partially known frame and introduce directly the complementary hypotheses into the frame itself. By example, if one knows only two hypotheses $\theta_1$, $\theta_2$ and their complements $\bar{\theta}_1$, $\bar{\theta}_2$, then can choose $\Theta=\{\theta_1,\theta_2,\bar{\theta}_1,\bar{\theta}_2\}$. In such case, we don't necessarily assume that $\bar{\theta}_1=\theta_2$ and $\bar{\theta}_2=\theta_1$ because $\bar{\theta}_1$ and $\bar{\theta}_2$ may include other unknown hypotheses we have no information about (case of partial known frame). More generally, in DSmT framework, it is not necessary that the frame is built on pure/simple (possibly vague) hypotheses $\theta_i$ as usually done in all theories managing uncertainty. The frame $\Theta$ can also contain directly as elements conjunctions and/or disjunctions (or mixed propositions) and negations/complements of pure hypotheses as well. The DSm rules also work in such non-classic frames because DSmT works on any distributive lattice built from $\Theta$ anywhere $\Theta$ is defined.

\subsection{Examples of combination rules}

Here are some numerical examples on results obtained by DSm rules of combination. More examples can be found in \cite{DSmTBook_2004a}.

\subsubsection{Example with $\Theta=\{\theta_1,\theta_2,\theta_3,\theta_4\}$}

Let's consider the frame of discernment $\Theta=\{\theta_1,\theta_2,\theta_3,\theta_4\}$, two independent experts, and the two following bbas
$$m_1(\theta_1)=0.6\quad m_1(\theta_3)=0.6\quad m_2(\theta_2)=0.6\quad m_2(\theta_4)=0.6$$
\noindent represented in terms of mass matrix 
\begin{equation*}
\mathbf{M}=
\begin{bmatrix}
0.6 & 0  & 0.4 & 0\\
0 & 0.2 & 0 & 0.8
\end{bmatrix}
\end{equation*}
\begin{itemize}
\item The Dempster's rule can not be applied because: $\forall 1\leq j \leq 4$, one gets $m(\theta_j) = 0/0$ (undefined!).
\item But the classic DSm rule works because one obtains: $m(\theta_1) = m(\theta_2) = m(\theta_3) = m(\theta_4) = 0$, 
and $m(\theta_1 \cap \theta_2) = 0.12$, $m(\theta_1 \cap \theta_4) = 0.48$,
$m(\theta_2 \cap \theta_3) = 0.08$, $m(\theta_3 \cap \theta_4) = 0.32$ (partial paradoxes/conflicts).
\item Suppose now one finds out that all intersections are empty (Shafer's model), then one applies the hybrid DSm rule and one gets (index $h$ stands here for {\it{hybrid}} rule): $m_h(\theta_1\cup \theta_2)=0.12$, $m_h(\theta_1\cup\theta_4)=0.48$, $m_h(\theta_2\cup\theta_3)=0.08$ and $m_h(\theta_3\cup\theta_4)=0.32$.
\end{itemize}

\subsubsection{Generalization of Zadeh's example with $\Theta=\{\theta_1,\theta_2,\theta_3\}$}

Let's consider $0 < \epsilon_1,\epsilon_2 < 1$ be two very tiny positive numbers (close to zero), the frame of discernment be $\Theta=\{\theta_1,\theta_2,\theta_3\}$, have two experts (independent sources of evidence $s_1$ and $s_2$) giving the belief masses
$$m_1(\theta_1)=1-\epsilon_1 \quad m_1(\theta_2)=0 \quad m_1(\theta_3)=\epsilon_1$$
$$m_2(\theta_1)=0 \quad m_2(\theta_2)=1-\epsilon_2 \quad m_2(\theta_3)=\epsilon_2$$
\noindent From now on, we prefer to use matrices to describe the masses, i.e.
$$\begin{bmatrix}
1-\epsilon_1 & 0 &\epsilon_1\\
0 & 1-\epsilon_2 & \epsilon_2
\end{bmatrix}
$$
\begin{itemize}
\item Using Dempster's rule of combination, one gets
$$m(\theta_3)=\frac{(\epsilon_1\epsilon_2)}{(1-\epsilon_1)\cdot 0 + 0\cdot (1-\epsilon_2) + \epsilon_1\epsilon_2}=1$$
\noindent which is absurd (or at least counter-intuitive). Note that whatever positive values for $\epsilon_1$, $\epsilon_2$ are, Dempster's rule of combination provides always the same result (one) which is abnormal. The only acceptable and correct result obtained by Dempster's rule is really obtained only in the trivial case when $\epsilon_1=\epsilon_2=1$, i.e. when both sources agree in $\theta_3$ with certainty which is obvious.
\item Using the DSm rule of combination based on free-DSm model, one gets 
$m(\theta_3)=\epsilon_1\epsilon_2$, $m(\theta_1\cap \theta_2)=(1-\epsilon_1)(1-\epsilon_2)$, $m(\theta_1\cap \theta_3)=(1-\epsilon_1)\epsilon_2$, $m(\theta_2\cap \theta_3)=(1-\epsilon_2)\epsilon_1$ and the others are zero which appears more reliable/trustable.
\item Going back to Shafer's model and using the hybrid DSm rule of combination, one gets 
$m(\theta_3)=\epsilon_1\epsilon_2$, $m(\theta_1\cup \theta_2)=(1-\epsilon_1)(1-\epsilon_2)$, $m(\theta_1\cup \theta_3)=(1-\epsilon_1)\epsilon_2$, $m(\theta_2\cup \theta_3)=(1-\epsilon_2)\epsilon_1$ and the others are zero.
\end{itemize}

\noindent
Note that in the special  case when $\epsilon_1=\epsilon_2=1/2$, one has
$$m_1(\theta_1)=1/2 \quad m_1(\theta_2)=0 \quad m_1(\theta_3)=1/2 \qquad\text{and}\qquad m_2(\theta_1)=0 \quad m_2(\theta_2)=1/2 \quad m_2(\theta_3)=1/2$$
Dempster's rule of combinations still yields $m(\theta_3)=1$ while the hybrid DSm rule based on the same Shafer's model yields now
$m(\theta_3)=1/4$, $m(\theta_1\cup \theta_2)=1/4$, $m(\theta_1\cup \theta_3)=1/4$, $m(\theta_2\cup \theta_3)=1/4$ which is normal.

\subsubsection{Comparison with Smets, Yager and Dubois \& Prade rules}

We compare the results provided by DSmT rules and the main common rules of combination on the following very simple numerical example where only 2 independent sources (a priori assumed equally reliable) are involved and providing their belief initially on the 3D frame $\Theta=\{\theta_1,\theta_2,\theta_3\}$. It is assumed in this example that Shafer's model holds and thus the belief assignments $m_1(.)$ and $m_2(.)$ do not  commit belief to internal conflicting information. $m_1(.)$ and $m_2(.)$ are chosen as follows:
 $$m_1(\theta_1)=0.1 \qquad m_1(\theta_2)=0.4 \qquad m_1(\theta_3)=0.2 \qquad m_1(\theta_1\cup \theta_2)=0.1$$
 $$m_2(\theta_1)=0.5 \qquad m_2(\theta_2)=0.1 \qquad m_2(\theta_3)=0.3 \qquad m_2(\theta_1\cup \theta_2)=0.1$$
 
\noindent
These belief masses are usually represented in the form of a belief mass matrix $\mathbf{M}$ given by
\begin{equation}
\mathbf{M}=
\begin{bmatrix}
0.1 & 0.4 & 0.2 & 0.3\\
0.5 & 0.1 & 0.3 & 0.1
\end{bmatrix}
\end{equation}
\noindent
where index $i$ for the rows corresponds to the index of the source no. $i$ and the indexes $j$ for columns of $\mathbf{M}$ correspond to a given choice for enumerating the focal elements of all sources.
In this particular example, index $j=1$ corresponds to $\theta_1$, $j=2$ corresponds to $\theta_2$, $j=3$ corresponds to $\theta_3$ and  $j=4$ corresponds to $\theta_1\cup \theta_2$.\\

Now let's imagine that one finds out that $\theta_3$ is actually truly empty because some extra and certain knowledge on $\theta_3$ is received by the fusion center. As example, $\theta_1$, $\theta_2$ and $\theta_3$ may correspond to three suspects (potential murders) in a police investigation, $m_1(.)$ and $m_2(.)$ corresponds to two reports of independent witnesses, but it turns out that finally $\theta_3$ has provided a strong alibi to the criminal police investigator once arrested by the policemen. This situation corresponds to set up a hybrid model $\mathcal{M}$ with the constraint $\theta_3\overset{\mathcal{M}}{=}\emptyset$. \\

Let's examine the result of the fusion in such situation obtained by the Smets', Yager's, Dubois \& Prade's and hybrid DSm rules of combinations. First note that, based on the free DSm model, one would get by applying the classic DSm rule (denoted here by index $DSmc$) the following fusion result

\begin{align*}
m_{DSmc}(\theta_1)&=0.21 \qquad m_{DSmc}(\theta_2)=0.11 \qquad m_{DSmc}(\theta_3)=0.06 \qquad m_{DSmc}(\theta_1\cup\theta_2)=0.03\\
m_{DSmc}(\theta_1\cap\theta_2)&=0.21 \qquad m_{DSmc}(\theta_1\cap\theta_3)=0.13\qquad m_{DSmc}(\theta_2\cap\theta_3)=0.14\\
m_{DSmc}(\theta_3\cap(\theta_1\cup\theta_2))&=0.11\\
\end{align*}

But because of the exclusivity constraints (imposed here by the use of Shafer's model and by the non-existential constraint $\theta_3\overset{\mathcal{M}}{=}\emptyset$), the total conflicting mass is actually given by
$$k_{12}=0.06 + 0.21 + 0.13 + 0.14 + 0.11=0.65 \qquad\text{(conflicting mass)}$$
\begin{itemize}
\item If one applies  {\bf{Dempster's rule}} \cite{Shafer_1976} (denoted here by index $DS$), one gets:
\begin{align*}
m_{DS}(\emptyset)& = 0\\
m_{DS}(\theta_1)&=0.21/[1- k_{12}]=0.21/[1-0.65]=0.21/0.35=0.600000\\
m_{DS}(\theta_2)&=0.11/[1-k_{12}]=0.11/[1-0.65]=0.11/0.35=0.314286\\
m_{DS}(\theta_1\cup\theta_2)&=0.03/[1-k_{12}]=0.03/[1-0.65]=0.03/0.35=0.085714
\end{align*}
\item If one applies {\bf{Smets' rule}} \cite{Smets_1994,Smets_2000} (i.e. the non normalized version of Dempster's rule with the conflicting mass transferred onto the empty set), one gets:
\begin{align*}
m_{S}(\emptyset)&=m(\emptyset)=0.65 \qquad\text{(conflicting mass)}\\
m_{S}(\theta_1)&=0.21\\
m_{S}(\theta_2)&=0.11\\
m_{S}(\theta_1\cup\theta_2)&=0.03
\end{align*}
\end{itemize}
\begin{itemize}
\item If one applies {\bf{Yager's rule}} \cite{Yager_1983,Yager_1985,Yager_1987}, one gets:
\begin{align*}
m_{Y}(\emptyset)&= 0\\
m_{Y}(\theta_1)&=0.21\\
m_{Y}(\theta_2)&=0.11\\
m_{Y}(\theta_1\cup\theta_2)&=0.03 + k_{12}=0.03+0.65=0.68
\end{align*}
\item If one applies  {\bf{Dubois \& Prade's rule}} \cite{Dubois_1988}, one gets because $\theta_3\overset{\mathcal{M}}{=}\emptyset$ :
\begin{align*}
m_{DP}(\emptyset)& = 0 \qquad \text{(by definition of Dubois \& Prade's rule)}\\
m_{DP}(\theta_1)&= [m_1(\theta_1)m_2(\theta_1) + m_1(\theta_1)m_2(\theta_1\cup\theta_2)+ m_2(\theta_1)m_1(\theta_1\cup\theta_2)]\\
& \quad + [m_1(\theta_1)m_2(\theta_3) + m_2(\theta_1)m_1(\theta_3)]\\
& = [0.1\cdot 0.5+0.1\cdot 0.1 +0.5\cdot 0.3] + [0.1\cdot 0.3 +0.5\cdot 0.2] = 0.21 + 0.13 = 0.34\\
m_{DP}(\theta_2)&=[0.4\cdot 0.1 + 0.4\cdot 0.1 +0.1\cdot 0.3] + [0.4\cdot 0.3 + 0.1\cdot 0.2] = 0.11 + 0.14 = 0.25\\
m_{DP}(\theta_1\cup\theta_2)&=[m_1(\theta_1\cup\theta_2)m_2(\theta_1\cup\theta_2)] 
+ [m_1(\theta_1\cup\theta_2)m_2(\theta_3) + m_2(\theta_1\cup\theta_2)m_1(\theta_3)]\\
& \quad + [m_1(\theta_1)m_2(\theta_2) + m_2(\theta_1)m_1(\theta_2)]\\
& = [0.3 0.1 ] + [0.3  \cdot 0.3 + 0.1  \cdot 0.2 ] + [0.1 \cdot 0.1 + 0.5  \cdot 0.4] = [0.03] + [0.09+0.02] + [0.01 + 0.20]\\
& = 0.03 + 0.11 + 0.21 = 0.35
\end{align*}
Now if one adds up the masses, one gets $0+ 0.34+0.25+0.35=0.94$ which is less than 1. Therefore Dubois \& Prade's rule of combination does not work when a singleton, or an union of singletons, becomes empty (in a dynamic fusion problem). The products of such empty-element columns of the mass matrix $\mathbf{M}$ are lost; this problem is fixed in DSmT by the sum $S_2(.)$ in \eqref{eq:DSmHkBis1} which transfers these products to the total or partial ignorances.\\
\end{itemize}
In this particular example, using the hybrid DSm rule, one transfers the product of the empty-element $\theta_3$ column, $m_1(\theta_3)m_2(\theta_3)=0.2\cdot 0.3=0.06$, to $m_{DSmh}(\theta_1\cup\theta_2)$, which becomes equal to $0.35+0.06=0.41$.\\

\subsection{Fusion of imprecise beliefs}
\label{sec2.7}

In many fusion problems, it seems very difficult (if not impossible) to have precise sources of evidence generating precise basic belief assignments (especially when belief functions are provided by human experts), and a more flexible plausible and paradoxical  theory supporting imprecise information becomes necessary. In the previous sections, we presented the fusion of {\it{precise}} uncertain and conflicting/paradoxical generalized basic belief assignments (gbba) in the DSmT framework. We mean here by precise gbba, basic belief functions/masses $m(.)$ defined precisely on the hyper-power set  $D^\Theta$ where each mass $m(X)$, where $X$ belongs to $D^\Theta$, is represented by only one real number belonging to $[0,1]$ such that $\sum_{X\in D^\Theta}m(X)=1$. In this section, we present the DSm fusion rule for dealing with {\it{admissible imprecise generalized basic belief assignments}} $m^I(.)$ defined as real subunitary intervals of $[0,1]$, or even more general as real subunitary sets [i.e. 
sets, not necessarily intervals]. An imprecise belief assignment $m^I(.)$ over $D^\Theta$ is said admissible if and only if there exists for every $X\in D^\Theta$ at least one real number $m(X)\in m^I(X)$ such that $\sum_{X\in D^\Theta}m(X)=1$. The idea to work with imprecise belief structures represented by real subset intervals of $[0,1]$ is not new and has been investigated in \cite{Lamata_1994,Denoeux_1997,Denoeux_1999} and references therein. The proposed works available in the literature, upon our knowledge were limited only to sub-unitary interval combination in the framework of Transferable Belief Model (TBM) developed by Smets \cite{Smets_1994,Smets_2000}. We extend the approach of Lamata \& Moral and Den\oe ux based on subunitary interval-valued masses to subunitary set-valued masses; therefore the 
closed intervals used by Den\oe ux to denote imprecise masses are generalized to any sets included in [0,1], i.e. in our case these sets can be unions of (closed, open, or half-open/half-closed) intervals and/or scalars all in $[0,1]$. Here, the proposed extension is done in the context of the DSmT framework, although it can also apply directly to fusion of imprecise belief structures within TBM as well if the user prefers to adopt TBM rather than DSmT.\\

Before presenting the general formula for the combination of generalized imprecise belief structures, we remind the following set operators involved in the formula. Several numerical examples are given in \cite{DSmTBook_2004a}.

\begin{itemize}
 \item
 {\bf{Addition of sets}}
 \begin{equation*}
 S_{1}\boxplus S_{2} =S_{2}\boxplus S_{1}\triangleq \{ x \mid x = s_{1}+s_{2},  s_{1} \in 
S_{1},s_{2} \in S_{2} \}
 \quad\text{with} \quad 
 \begin{cases}
\inf(S_{1}\boxplus S_{2})=\inf(S_1) + \inf(S_2)\\
\sup(S_{1}\boxplus S_{2})=\sup (S_1) + \sup(S_2)
\end{cases}
\label{eq:addition}
\end{equation*}
\item
{\bf{Subtraction of sets}}
 \begin{equation*}
 S_{1}\boxminus S_{2} \triangleq \{ x \mid x = s_{1}-s_{2},  s_{1} \in 
S_{1}, s_{2} \in S_{2} \}
 \quad\text{with} \quad 
\begin{cases}\inf(S_{1}\boxminus S_{2})=\inf(S_1) - \sup(S_2)\\
\sup(S_{1}\boxminus S_{2})=\sup(S_1) - \inf(S_2)
\end{cases}
\label{eq:addition}
 \end{equation*}
  \item
 {\bf{Multiplication of sets}}
 \begin{equation*}
S_{1}\boxdot S_{2} \triangleq \{ x \mid x = s_{1}\cdot s_{2}, s_{1} \in 
S_{1},s_{2} \in S_{2} \}
 \quad\text{with} \quad 
\begin{cases}
\inf(S_{1}\boxdot S_{2})=\inf(S_{1})\cdot \inf(S_{2})\\
\sup(S_{1}\boxdot S_{2})=\sup(S_{1})\cdot \sup (S_{2})
\end{cases}
\label{eq:multiplication}
 \end{equation*}
\end{itemize}

\subsubsection{DSm rule of combination for imprecise beliefs}
We present the generalization of the DSm rules to combine any type of imprecise belief assignment which may be represented by the union of several sub-unitary (half-) open intervals, (half-)closed intervals and/or sets of points belonging to [0,1]. Several numerical examples are also given. In the sequel, one uses the notation $(a,b)$ for an open interval, $[a,b]$ for a closed interval, and $(a,b]$ or $[a,b)$ for a half open and half closed interval. From the previous operators on sets, one can generalize the DSm rules (classic and hybrid) from scalars to sets in the following way \cite{DSmTBook_2004a} (chap. 6): $\forall A\neq\emptyset \in D^\Theta$,
\begin{equation}
m^I(A) = \underset{\underset{(X_1\cap X_2\cap\ldots\cap X_k)=A}{X_1,X_2,\ldots,X_k\in D^\Theta}}{\boxed{\sum}}
\underset{i=1,\ldots,k}{\boxed{\prod}} m_i^I(X_i)
\label{eq:DSMruleSetsImprecise}
\end{equation}
\noindent
where $\boxed{\sum}$ and $\boxed{\prod}$ represent the summation, and respectively product, of sets.\\

Similarly, one can generalize the hybrid DSm rule from scalars to sets in the following way:
\begin{equation}
m_{\mathcal{M}(\Theta)}^I(A)\triangleq 
\phi(A)\boxdot \Bigl[ S_1^I(A) \boxplus S_2^I(A) \boxplus S_3^I(A)\Bigr]
 \label{eq:DSmHkBisImprecise}
\end{equation}
\noindent
$\phi(A)$ is the {\it{characteristic non emptiness function}} of the set $A$ and $S_1^I(A)$, $S_2^I(A)$ and $S_3^I(A)$ are defined by
\begin{equation}
S_1^I(A)\triangleq
\underset{\underset{(X_1\cap X_2\cap\ldots\cap X_k)=A}{X_1,X_2,\ldots,X_k\in D^\Theta}}{\boxed{\sum}}
\underset{i=1,\ldots,k}{\boxed{\prod}} m_i^I(X_i)
\label{eq:S1I}
\end{equation}
\begin{equation}
S_2^I(A)\triangleq 
\underset{\underset{[\mathcal{U}=A]\vee [(\mathcal{U}\in\boldsymbol{\emptyset}) \wedge (A=I_t)]}{X_1,X_2,\ldots,X_k\in\boldsymbol{\emptyset}}}{\boxed{\sum}}
\underset{i=1,\ldots,k}{\boxed{\prod}} m_i^I(X_i)
\label{eq:S2I}
\end{equation}
\begin{equation}
S_3^I(A)\triangleq
\underset{\underset{(X_1\cap X_2\cap\ldots\cap X_k)\in\boldsymbol{\emptyset} }{\underset{(X_1\cup X_2\cup\ldots\cup X_k)=A}{X_1,X_2,\ldots,X_k\in D^\Theta}}}{\boxed{\sum}}
\underset{i=1,\ldots,k}{\boxed{\prod}} m_i^I(X_i)
\label{eq:S3I}
\end{equation}
In the case when all sets are reduced to points (numbers), the set operations become normal operations with numbers; the sets operations are generalizations of numerical operations. When imprecise belief structures reduce to precise belief structure, DSm rules  \eqref{eq:DSMruleSetsImprecise} and \eqref{eq:DSmHkBisImprecise} reduce to their precise version  \eqref{JDZT}  and  \eqref{eq:DSmHkBis1} respectively.

\subsubsection{Example}

Here is a simple example of fusion with with multiple-interval masses. For simplicity, this example is a particular case when the theorem of admissibility (see \cite{DSmTBook_2004a} p. 138 for details) is verified by a few points, which happen to be just on the bounders. It is an extreme example, because we tried to comprise all kinds of possibilities which may occur in the imprecise or very imprecise fusion. So, let's consider a fusion problem over $\Theta=\{\theta_1,\theta_2\}$, two independent sources of information with the following imprecise admissible belief assignments
\begin{table}[h]
\begin{equation*}
\begin{array}{|c|c|c|}
\hline
A\in D^\Theta & m_1^I(A) & m_2^I(A) \\
\hline
\theta_1 & [0.1,0.2] \cup \{0.3\} & [0.4,0.5]\\
\theta_2 &(0.4,0.6)\cup [0.7,0.8] &  [0,0.4]\cup \{0.5,0.6\}\\
\hline
\end{array}
\end{equation*}
\caption{Inputs of the fusion with imprecise bba}
\label{mytablex1}
\end{table}

\noindent
Using the DSm classic rule for sets, one gets
\begin{equation*}
m^I(\theta_1)=([0.1,0.2] \cup \{0.3\})\boxdot [0.4,0.5] = ([0.1,0.2] \boxdot [0.4,0.5])\cup (\{0.3\}\boxdot [0.4,0.5] )= [0.04,0.10] \cup [0.12,0.15]
\end{equation*}
\begin{align*}
m^I(\theta_2)& =((0.4,0.6)\cup [0.7,0.8] )\boxdot ([0,0.4]\cup \{0.5,0.6\})\\
&= ((0.4,0.6)\boxdot [0,0.4])\cup ((0.4,0.6)\boxdot  \{0.5,0.6\}) \cup ([0.7,0.8]\boxdot [0,0.4]) \cup ([0.7,0.8]\boxdot \{0.5,0.6\})\\
&= (0,0.24)\cup (0.20,0.30) \cup (0.24,0.36)\cup [0,0.32] \cup [0.35,0.40] \cup [0.42,0.48] = [0,0.40] \cup [0.42,0.48]
\end{align*}
\begin{align*}
m^I(\theta_1\cap \theta_2)&=
[([0.1,0.2] \cup \{0.3\})\boxdot([0,0.4]\cup \{0.5,0.6\})] \boxplus [[0.4,0.5]\boxdot ((0.4,0.6)\cup [0.7,0.8]) ]\\
&=[ ([0.1,0.2]\boxdot [0,0.4]) \cup ([0.1,0.2]\boxdot \{0.5,0.6\}) \cup ( \{0.3\}\boxdot [0,0.4]) \cup ( \{0.3\}\boxdot  \{0.5,0.6\})] \\
& \quad \boxplus [ ([0.4,0.5]\boxdot (0.4,0.6)) \cup ([0.4,0.5]\boxdot [0.7,0.8] ) ] \\
& = [[0,0.08]\cup [0.05,0.10]\cup [0.06,0.12] \cup [0,0.12] \cup \{0.15,0.18\}] \boxplus [(0.16,0.30)\cup[0.28,0.40]]\\
&= [[0,0.12]\cup \{0.15,0.18\}]\boxplus (0.16,0.40] =(0.16,0.52] \cup (0.31,0.55] \cup (0.34,0.58]=(0.16,0.58]
\end{align*}
\noindent
Hence finally the fusion admissible result is given by:
\begin{table}[h]
\begin{equation*}
\begin{array}{|c|c|}
\hline
A\in D^\Theta & m^I(A)= [m_1^I \oplus m_2^I](A) \\
\hline
\theta_1 & [0.04,0.10] \cup [0.12,0.15]\\
\theta_2 & [0,0.40] \cup [0.42,0.48] \\
\theta_1\cap \theta_2  & (0.16,0.58]\\
\theta_1\cup \theta_2 & 0 \\
\hline
\end{array}
\end{equation*}
\caption{Fusion result with the DSm classic rule}
\label{mytablex2}
\end{table}

\noindent
If one finds out\footnote{We consider now a dynamic fusion problem.} that $\theta_1\cap \theta_2 \overset{\mathcal{M}}{\equiv}\emptyset$ (this is our hybrid model $\mathcal{M}$ one wants to deal with), then one uses the hybrid DSm rule for sets \eqref{eq:DSmHkBisImprecise}: $m_{\mathcal{M}}^I(\theta_1\cap \theta_2)=0$ and $m_{\mathcal{M}}^I(\theta_1\cup \theta_2)= (0.16,0.58]$, the others imprecise masses are not changed. In other words, one gets now with hybrid DSm rule applied to imprecise beliefs:

\begin{table}[h]
\begin{equation*}
\begin{array}{|c|c|}
\hline
A\in D^\Theta & m_{\mathcal{M}}^I(A)= [m_1^I \oplus m_2^I](A) \\
\hline
\theta_1 & [0.04,0.10] \cup [0.12,0.15]\\
\theta_2 & [0,0.40] \cup [0.42,0.48] \\
\theta_1\cap \theta_2\overset{\mathcal{M}}{\equiv}\emptyset  & 0 \\
\theta_1\cup \theta_2 & (0.16,0.58]\\
\hline
\end{array}
\end{equation*}
\caption{Fusion result with the hybrid DSm rule for $\mathcal{M}$ }
\label{mytablex3}
\end{table}

Let's check now the admissibility conditions and theorem. For the source 1, there exist the precise masses $(m_1(\theta_1)=0.3) \in ([0.1,0.2] \cup \{0.3\})$ and $(m_1(\theta_2)=0.7) \in ((0.4,0.6)\cup [0.7,0.8])$ such that $0.3+0.7=1$. For the source 2, there exist the precise masses $(m_1(\theta_1)=0.4) \in ([0.4,0.5])$ and $(m_2(\theta_2)=0.6) \in ([0,0.4]\cup \{0.5,0.6\})$ such that $0.4+0.6=1$. Therefore both sources associated with $m_1^I(.)$ and $m_2^I(.)$ are admissible imprecise sources of information.\\

It can be easily checked that the DSm classic fusion of $m_1(.)$ and $m_2(.)$ yields the paradoxical basic belief assignment $m(\theta_1)=[m_1\oplus m_2](\theta_1)=0.12$, $m(\theta_2)=[m_1\oplus m_2](\theta_2)=0.42$ and $m(\theta_1\cap \theta_2)=[m_1\oplus m_2](\theta_1\cap \theta_2)=0.46$.
One sees that the admissibility theorem is satisfied since $(m(\theta_1)=0.12)\in (m^I(\theta_1)=[0.04,0.10] \cup [0.12,0.15])$, $(m(\theta_2)=0.42)\in (m^I(\theta_2)=[0,0.40] \cup [0.42,0.48])$ and $(m(\theta_1\cap \theta_2)=0.46)\in (m^I(\theta_1\cap \theta_2)=(0.16,0.58])$ such that $0.12+0.42+0.46=1$. Similarly if one finds out that $\theta_1\cap\theta_2=\emptyset$, then one uses the hybrid DSm rule and one gets: $m(\theta_1\cap\theta_2)=0$ and $m(\theta_1\cup\theta_2)=0.46$; the others remain unchanged. The admissibility theorem still holds, because one can pick at least one number in each subset $m^I(.)$ such that the sum of these numbers is  1. This approach can be also used in the similar manner to obtain imprecise pignistic probabilities from $m^I(.)$ for decision-making under uncertain, paradoxical and imprecise sources of information as well. The generalized pignistic transformation (GPT) is presented in next section.

\subsection{The generalized pignistic transformation (GPT)}

\subsubsection{The classical pignistic transformation}

We follow here the Smets'  vision which considers the management of information as a two 2-levels process: credal (for combination of evidences) and pignistic\footnote{Pignistic terminology has been coined by Philippe Smets and comes from {\it{pignus}}, a bet in Latin.} (for decision-making) , i.e "{\it{when someone must take a decision, he must then construct a probability function derived from the belief function that describes his credal state. This probability function is then used to make decisions}}" \cite{Smets_1988} (p. 284). One obvious way to build this probability function corresponds to the so-called Classical Pignistic Transformation (CPT) defined in the DST framework (i.e. based on the Shafer's model assumption) as \cite{Smets_2000}:

\begin{equation}
P\{A\}=\sum_{X \in 2^\Theta}\frac{|X\cap A|}{|X|}m(X)
\label{eq:Pig}
\end{equation}

where $|A|$ denotes the number of worlds in the set $A$ (with convention $|\emptyset | / |\emptyset |=1$, to define $P\{\emptyset \}$). $P\{A\}$ corresponds to $BetP(A)$ in Smets' notation \cite{Smets_2000}. Decisions are achieved by computing the expected utilities of the acts using the subjective/pignistic $P\{.\}$ as the probability function needed to compute expectations.
Usually, one uses the maximum of the pignistic probability as decision criterion. The max. of $P\{.\}$ is often considered as a prudent betting decision criterion between the two other alternatives (max of plausibility or max. of credibility which appears to be respectively too optimistic or too pessimistic). It is easy to show that $P\{.\}$ is indeed a probability function (see \cite{Smets_1994}).

\subsubsection{Notion of DSm cardinality}
One important notion involved in the definition of the Generalized Pignistic Transformation (GPT) is the {\it{DSm cardinality}}. The {\it{DSm cardinality}} of any element $A$ of hyper-power set  $D^\Theta$, denoted $\mathcal{C}_\mathcal{M}(A)$, corresponds to the number of parts of $A$ in the corresponding fuzzy/vague Venn diagram of the problem (model $\mathcal{M}$) taking into account the set of integrity constraints (if any), i.e. all the possible intersections due to the nature of the elements $\theta_i$. This {\it{intrinsic cardinality}} depends on the model $\mathcal{M}$ (free, hybrid or Shafer's model).  $\mathcal{M}$ is the model that contains $A$, which depends both on the
dimension $n=\vert \Theta \vert$ and on the number of non-empty intersections present in its associated Venn diagram (see \cite{DSmTBook_2004a} for details ). The DSm cardinality depends on the cardinal of $\Theta = \{\theta_1,\theta_2,\ldots,\theta_n\}$ and on the model of $D^\Theta$ (i.e., the number of intersections and between what elements of $\Theta$ - in a word the structure) at the same time; it is not necessarily that every singleton, say $\theta_i$, has the same DSm cardinal, because each singleton has a different structure; if its structure is the simplest (no intersection of this elements with other elements) then $\mathcal{C}_\mathcal{M}(\theta_i)=1$, if the structure is more complicated (many intersections) then $\mathcal{C}_\mathcal{M}(\theta_i) > 1$; let's consider a singleton $\theta_i$: if it has 1 intersection only then $\mathcal{C}_\mathcal{M}(\theta_i)=2$, for 2 intersections only $\mathcal{C}_\mathcal{M}(\theta_i)$ is 3 or 4 depending on the model $\mathcal{M}$, for $m$ intersections it is between $m+1$ and $2^m$ depending on the model; the maximum DSm cardinality is $2^{n-1}$ and occurs for $\theta_1\cup\theta_2\cup\ldots\cup\theta_n$ in the free model $\mathcal{M}^f$; similarly for any set from $D^\Theta$: the more complicated structure it has, the bigger is the DSm cardinal;
thus the DSm cardinality measures the complexity of en element from $D^\Theta$, which is a nice characterization in our opinion; we may say that for the singleton $\theta_i$ not even $\vert\Theta\vert$ counts, but only its structure (= how many other singletons intersect $\theta_i$). Simple illustrative examples are given in Chapter 3 and 7 of \cite{DSmTBook_2004a}.
One has $1 \leq \mathcal{C}_\mathcal{M}(A) \leq 2^n-1$. $\mathcal{C}_\mathcal{M}(A)$ must not be confused with the classical cardinality $\vert A \vert$ of a given set $A$ (i.e. the number of its distinct elements) - that's why a new notation is necessary here.
$\mathcal{C}_\mathcal{M}(A)$ is very easy to compute by programming from the algorithm of generation of $D^\Theta$ given explicated in \cite{DSmTBook_2004a}.

As example, let's take back the example of the simple hybrid DSm model described in section \ref{Sec:DSMmodels}, then one gets the following list of elements (with their DSm cardinal) for the restricted $D^\Theta$ taking into account the integrity constraints of this hybrid model:
\begin{equation*}
\begin{array}{lcl}
A\in D^\Theta                     & \mathcal{C}_{\mathcal{M}}(A) \\
\hline
\alpha_0\triangleq\emptyset                                                 & 0 \\
\alpha_1\triangleq\theta_1\cap\theta_2                             & 1 \\
\alpha_2\triangleq\theta_3                                                    & 1 \\
\alpha_3\triangleq\theta_1                                                    & 2 \\
\alpha_4\triangleq\theta_2                                                    & 2 \\
\alpha_5\triangleq\theta_1\cup\theta_2                             & 3 \\
\alpha_6\triangleq\theta_1\cup\theta_3                             & 3 \\
\alpha_7\triangleq\theta_2\cup\theta_3                             & 3 \\
\alpha_8\triangleq\theta_1\cup\theta_2\cup\theta_3       & 4  \\
\end{array}
\end{equation*}
\begin{center}
{\bf{Eaxmple of DSm cardinals}}: $\mathcal{C}_{\mathcal{M}}(A)$ for hybrid model $\mathcal{M}$
\end{center}

\subsubsection{The  Generalized Pignistic Transformation}

To take a rational decision within the DSmT framework, it is necessary to generalize the Classical Pignistic Transformation in order to construct a pignistic probability function from any generalized basic belief assignment $m(.)$ drawn from the DSm rules of combination. Here is the simplest and direct extension of the CPT to define the Generalized Pignistic Transformation:
\begin{equation}
\forall A \in D^\Theta, \qquad \qquad P\{A\}=\sum_{X \in D^\Theta}  \frac{\mathcal{C}_{\mathcal{M}}(X\cap A)}{\mathcal{C}_{\mathcal{M}}(X)}m(X)
\label{eq:PigG}
\end{equation}
\noindent
where $\mathcal{C}_{\mathcal{M}}(X)$ denotes the DSm cardinal of proposition $X$ for the DSm model $\mathcal{M}$ of the problem under consideration.\\

The decision about the solution of the problem  is usually taken by the maximum of pignistic probability function $P\{.\}$. Let's remark the close ressemblance of the two pignistic transformations \eqref{eq:Pig} and \eqref{eq:PigG}. It can be shown that \eqref{eq:PigG} reduces to \eqref{eq:Pig} when the hyper-power set $D^\Theta$ reduces to classical power set $2^\Theta$ if we adopt Shafer's model. But  \eqref{eq:PigG} is a generalization of  \eqref{eq:Pig} since it can be used for computing pignistic probabilities for any models (including Shafer's model). It has been proved in \cite{DSmTBook_2004a} (Chap. 7) that $P\{.\}$ is indeed a probability function.

\section{Fuzzy Inference for Information Fusion}

We further connect the fusion rules of combination with fuzzy and neutrosophic operators.  LetÕs first replace the Conjunctive Rule and Disjunctive Rule with the fuzzy T-norm and T-conorm versions respectively. These rules started from the T-norm and T-conorm respectively in fuzzy and neutrosophic logics, where the {\it{and}} logic operator $\wedge$ corresponds in fusion to the conjunctive rule, while the {\it{or}} logic operator $\vee$ corresponds to the disjunctive rule.
While the logic operators deal with degrees of truth and degrees of falsehood, the fusion rules deal with degrees of belief and degrees of disbelief of hypotheses.

\subsection{T-Norm}
A T-norm is a function $T_n: {[0,1]}^2 \mapsto  [0,1]$, defined in fuzzy set theory and fuzzy logic to represent the {\it{intersection}} of two fuzzy sets and the fuzzy logical operator {\it{and}} respectively.  Extended to the fusion theory the T-norm will be a substitute for the conjunctive rule.
The T-norm satisfies the conditions:
\begin{itemize}
 \item[a)] Boundary Conditions:   $T_n(0, 0) = 0, T_n(x, 1) = x$
 \item[b)] Commutativity: $T_n(x, y) = T_n(y, x)$       
 \item[c)] Monotonicity: If  $x \leq u$ and $y \leq v$, then $T_n(x, y)\leq T_n(u, v)$
 \item[d)] Associativity: $T_n(T_n(x, y), z ) = T_n( x, T_n(y, z) )$        
\end{itemize}
\noindent
There are many functions which satisfy the T-norm conditions.  We present below the most known ones:
\begin{itemize}
\item The Algebraic Product T-norm:  $T_{\text{n-algebraic}}(x, y) = x\cdot y$
\item The Bounded T-norm: $T_{\text{n-bounded}}(x, y) = \max \{0, x+y-1\}$
\item The Default (min) T-norm (introduced by Zadeh): $T_{\text{n-min}}(x, y) = \min \{x, y\}$
\end{itemize}

\subsection{T-conorm}

A T-conorm is a function $T_c: {[0,1]}^2 \mapsto [0, 1]$, defined in fuzzy set theory and fuzzy logic to represent the {\it{union}} of two fuzzy sets and the fuzzy logical operator {\it{or}} respectively.  Extended to the fusion theory the T-conorm will be a substitute for the disjunctive rule.
The T-conorm satisfies the conditions:
\begin{itemize}
 \item[a)] Boundary Conditions:   $T_c(1, 1) = 1, T_c(x, 0) = x$
 \item[b)] Commutativity: $T_c(x, y) = T_c(y, x)$        
 \item[c)] Monotonicity: if  $x \leq u$ and $y \leq v$, then $T_c(x, y)\leq T_c(u, v)$ 
 \item[d)] Associativity: $T_c(T_c(x, y), z ) = T_c( x, T_c(y, z) )$ 
\end{itemize}
\noindent
       
There are many functions which satisfy the T-conorm conditions.  We present below the most known ones:
\begin{itemize}
\item The Algebraic Product T-conorm: $T_{\text{c-algebraic}}(x, y) = x+y-x\cdot y$
\item The Bounded T-conorm: $T_{\text{c-bounded}}(x, y) = \min \{1, x+y\}$
\item The Default (max) T-conorm (introduced by Zadeh): $T_{\text{c-max}}(x, y) = \max \{x, y\}$
\end{itemize}

Then, the {\it{T-norm Fusion rule}} is defined as follows: $m_{\cap 12}(A) = \sum_{\substack{Y,Y\in \Theta\\ X\cap Y = A}}  T_n(m_1(X),m_2(Y))$
and the {\it{T-conorm Fusion rule}} is defined as follows:
$m_{\cup 12}(A) = \sum_{\substack{Y,Y\in \Theta\\ X\cup Y = A}}  T_c(m_1(X),m_2(Y))$.\\

The min T-norm rule yields results, very closed to Conjunctive Rule. It satisfies the principle of neutrality of the vacuous bba, reflects the majority opinion, converges towards idempotence.  It is simpler to apply, but needs normalization. What is missed it is a strong justification of the way of presenting the fusion process.  But we think, the consideration between two sources of information as a vague relation, characterized with the particular way of association between focal elements, and corresponding degree of association (interaction) between them is reasonable. Min rule can be interpreted as an optimistic lower bound for combination of bba and the below Max rule as a prudent/pessimistic upper bound. The T-norm and T-conorm are commutative, associative, isotone, and have a neutral element.

\section{Degrees of intersection, union, inclusion}

In order to improve many fusion rules we can insert a degree of intersection, a degree of union, or a degree of inclusion.  These are defined as follows:

\subsection{Degree of Intersection}

The degree of intersection measures the percentage of overlapping region of two sets $X_1$, $X_2$ with respect to the whole reunited regions of the sets using the cardinal of sets not the fuzzy set point of view:
$$d(X_1\cap X_2) = \frac{|X_1\cap X_2|}{|X_1\cup X_2|}$$
\noindent
where $|X|$ means cardinal of the set $X$. \\

For the minimum intersection/overlapping, i.e. when $X_1\cap X_2=\emptyset$, the degree of intersection is 0, while for the maximum intersection/overlapping, i.e. when $X_1 = X_2$, the degree of intersection is 1.

\subsection{Degree of Union}
The degree of intersection measures the percentage of non-overlapping region of two sets $X_1$, $X_2$ with respect to the whole reunited regions of the sets using the cardinal of sets not the fuzzy set point of view:
$$d(X_1\cup X_2) = \frac{|X_1\cup X_2|-|X_1\cap X_2|}{|X_1\cup X_2|}$$
For the maximum non-overlapping, i.e. when $X_1\cap X_2=\emptyset$, the degree of union is 1, while for the minimum non-overlapping, i.e. when $X_1 = X_2$, the degree of union is 0. The sum of degrees of intersection and union is 1 since they complement each other.

\subsection{Degree of inclusion}

The degree of inclusion measures the percentage of the included region $X_1$ with respect to the includant region $X_2$: Let $X_1 \subseteq X_2$, then
     $$ d(X_1 \subseteq X_2) =\frac{|X_1|}{|X_2|}$$

\noindent
$d(\emptyset \subseteq X_2) = 0$ because nothing (i.e. empty set) is included in $X_2$, while $d(X_2 \subseteq X_2) = 1$ because $X_2$ is fulfilled by inclusion. By definition $d(\emptyset \subseteq\emptyset) = 1$. We can generalize the above degree for $n \geq 2$ sets.    

\subsection{Improvements of belief and plausibility functions }

Thus the $\text{Bel}(.)$ and $\text{Pl}(.)$ functions can incorporate in their formulas the above degrees of inclusion and intersection respectively:
 
\begin{itemize}
\item Belief function improved: $\forall A\in D^\Theta\setminus\emptyset, \text{Bel}_d(A)=
\sum_{\substack{X\in D^\Theta\\ X\subseteq A}}\frac{|X|}{|A|}m(X)$
\item Plausibility function improved: $\forall A\in D^\Theta\setminus\emptyset, \text{Pl}_d(A)=
\sum_{\substack{X\in D^\Theta\\ X\cap A\neq\emptyset}}\frac{|X\cap A|}{|X\cup A|}m(X)$
\end{itemize} 

 \subsection{Improvements of fusion rules}

\begin{itemize}
\item Disjunctive rule improved:
$$\forall A\in D^\Theta\setminus\emptyset, \qquad m_{\cup d}(A)=
k_{\cup d}\cdot \sum_{\substack{X_1,X_2\in D^\Theta\\ X_1\cup X_2 =A}}
\frac{|X_1\cup X_2|-|X_1\cap X_2|}{|X_1\cup X_2|}m_1(X_1)m_2(X_2)$$
\noindent
where $k_{\cup d}$ is a constant of normalization.
\item Dezert-Smarandache classical rule improved:
$$\forall A\in D^\Theta\setminus\emptyset, \qquad m_{DSmCd}(A)=
k_{DSmCd}\cdot \sum_{\substack{X_1,X_2\in D^\Theta\\ X_1\cap X_2 =A}}
\frac{|X_1\cap X_2|}{|X_1\cup X_2|}m_1(X_1)m_2(X_2)$$
\noindent
where $k_{DSmCd}$ is a constant of normalization. This rule is similar with the ZhangÕs Center Combination rule \cite{Zhang_1994} extended on the Boolean algebra $(\Theta, \cup,\cap, \mathcal{C})$ and using another definition for the degree of intersection (here $ \mathcal{C}$ denotes the complement).
\item Dezert-Smarandache hybrid rule improved:
\begin{multline*}
\forall A\in D^\Theta\setminus\emptyset, \qquad m_{DSmHd}(A)=
k_{DSmCd}\cdot \{ \sum_{\substack{X_1,X_2\in D^\Theta\\ X_1\cap X_2 =A}}
\frac{|X_1\cap X_2|}{|X_1\cup X_2|}m_1(X_1)m_2(X_2) \\
+
\sum_{\substack{X_1,X_2\in \boldsymbol{\emptyset}\\ [\mathcal{U}=A]\vee [(\mathcal{U}\in\boldsymbol{\emptyset}) \wedge (A=I_t)]}}m_1(X_1)m_2(X_2)
+
\sum_{\substack{X_1,X_2\in D^\Theta \\ u(c(X_1\cap X_2))=A \\ (X_1\cap X_2)\in\boldsymbol{\emptyset}}}  \frac{|X_1\cup X_2|-|X_1\cap X_2|}{|X_1\cup X_2|}m_1(X_1)m_2(X_2)
\}
\end{multline*}
\noindent
where $k_{DSmHd}$ is a constant of normalization.
\end{itemize} 

\section{Neutrosophic Inference for Information Fusion}

Similarly to the fuzzy improvement of the fusion rules we can now consider the neutrosophic improvement of the fusion rules of combination. LetÕs now replace the Conjunctive Rule and Disjunctive Rule with the neutrosophic N-norm and N-conorm versions respectively \cite{Wang_2004}.

\subsection{Neutrosophy}

Neutrosophic Logic, Neutrosophic Set, and Neutrosophic Probability started from Neutrosophy \cite{Smarandache_2000,Smarandache_2002,Smarandache_2002a,Smarandache_2002b}. Neutrosophy is a new branch of philosophy which studies the origin, nature, and scope of neutralities, as well as their interactions with different ideational spectra. It is an extension of dialectics.  Its fundamental theory is that every idea $<A>$ tends to be neutralized, diminished, balanced by $<NonA>$ ideas (not only $<AntiA>$ as Hegel asserted) - as a state of equilibrium, where $<NonA> = \text{what is not} <A>$, $<AntiA> = \text{the opposite of} <A>$, and $<NeutA> = \text{what is neither} <A> \text{nor} <AntiA>$.
\subsection{Nonstandard analysis}

\subsubsection{Short introduction}
Abraham Robinson developed the nonstandard analysis in sixties \cite{Robinson_1966}.
$x$ is called {\it{infinitesimal}} if $|x|<1/n$ for any positive $n$.
A {\it{left monad}} is defined by $(^-a) =\{a-x | x\in\mathbb{R}^{\star}, x>0 \, \text{infinitesimal} \} = a-\epsilon$ and a right monad by $(b^+)=\{b+x | x\in\mathbb{R}^{\star}, x>0 \, \text{infinitesimal} \} = b+\epsilon$ where $\epsilon >0$ is infinitesimal; $a$, $b$ are called {\it{standard parts}}, $\epsilon$ is called {\it{nonstandard}} part. A bimonad is defined as $(^-a^+) = (^-a)\cup(a^+)$.

\subsubsection{Operations with nonstandard finite real numbers}
$$^-a\star b = ^-(a\star b)\qquad a\star b^+ = (a\star b)^+\qquad  ^-a\star b^+ = ^-(a\star b)^+$$
\begin{itemize}
\item the left monads absorb themselves: $^-a\star ^-b = ^-(a\star b)$
\item the right monads absorb themselves: $a^+ \star b^+ = {(a\star b)}^+$
\end{itemize}
 \noindent
 where $\star$ operation can be addition, subtraction, multiplication, division and power. The operations with real standard or non-standard subsets are defined according definitions given in section \ref{sec2.7}.

\subsection{Neutrosophic logic}

LetÕs consider the nonstandard unit interval $]^-0, 1^+[$, with left and right borders vague, imprecise.  Let $T$, $I$, $F$ be standard or nonstandard subsets of $]^-0, 1^+[$. Then:
Neutrosophic Logic (NL) is a logic in which each proposition is $T\%$ true, $I\%$ indeterminate, and $F\%$ false, where:

$$-0 \leq \inf T + \inf I + \inf F \leq \sup T + \sup I + \sup F \leq 3^+$$

$T$, $I$, $F$  are not necessary intervals, but any sets (discrete, continuous, open or closed or half-open/half-closed interval, intersections or unions of the previous sets, etc.).\\
 
\noindent
 For example: proposition $P$ is between 30-40\% or 45-50\% true, 20\% indeterminate, and 60\% or between 66-70\% false (according to various analyzers or parameters). NL is a generalization of ZadehÕs fuzzy logic (FL), especially of AtanassovÕs intuitionistic fuzzy logic (IFL) \cite{Atanassov_1986,Atanassov_1990,Atanassov_1998}, and other logics.

\subsection{Differences between Neutrosophic Logic and Intuitionistic Fuzzy Logic}

\begin{itemize}
\item[a)] In NL there is no restriction on $T$, $I$, $F$, while in IFL the sum of components (or their superior limits) $= 1$;  thus NL can characterize the incomplete information (sum $< 1$), paraconsistent  information (sum $> 1$).  
\item[b)] NL can distinguish, in philosophy, between absolute truth [NL(absolute truth)$=1^+$] and relative truth [NL(relative truth)$=1$], while IFL cannot; {\it{absolute truth}} is truth in all possible worlds (Leibniz), {\it{relative truth}} is truth in at least one world.
\item[c)] In NL the components can be nonstandard, in IFL they donÕt.
\item[d)] NL, like {\it{dialetheism}} [some contradictions are true], can deal with paradoxes, NL(paradox) $= (1,I,1)$, while IFL cannot.
\end{itemize}

\subsection{Neutrosophic Logic generalizes many logics}

Let the components reduced to scalar numbers, $t$, $i$, $f$, with $t+i+f=n$; NL generalizes: 
\begin{itemize}
\item the {\it{Boolean logic}} (for $n = 1$ and $i= 0$, with $f$, $f$ either 0 or 1);
\item the {\it{multi-valued logic}}, which supports the existence of many values between true and false - Lukasiewicz, 3 values \cite{Lukasiewicz_1904,Lukasiewicz_1910}; Post, $m$ values -  (for $n = 1$, $Ii= 0$, $0\leq t, f \leq 1$);
\item  the {\it{intuitionistic logic}}, which supports incomplete theories, where $A\wedge \neg A$ not always true, and $\exists xP(x)$ needs an algorithm constructing $x$ \cite{Brouwer_1907,Brouwer_1975,Brouwer_1981,Brouwer_1992,Heyting_1965} (for $0 < n < 1$ and $i = 0$, $0\leq t, f \leq 1$);
\item the {\it{fuzzy logic}}, which supports degrees of truth \cite{Zadeh_1965} (for $n = 1$ and $i = 0$, $0\leq t, f \leq 1$);
\item  the {\it{intuitionistic fuzzy logic}}, which supports degrees of truth and degrees of falsity while whatÕs left is considered indeterminacy \cite{Atanassov_1990} (for $n = 1$);
\item  the {\it{paraconsistent logic}}, which supports conflicting information, and Ôanything follows from contradictionsÕ fails, i.e. $\neg A \wedge A \supset B$ fails;  $\neg A \wedge A$ is not always false (for $n > 1$ and $i= 0$, with both $0< t$, $f < 1$); 
\item the {\it{dialetheism}}, which says that some contradictions are true, $\neg A \wedge A=\text{true}$ (for $t= f = 1$ and $i = 0$; some paradoxes can be denoted this way too);
\item	 the {\it{faillibilism}}, which says that uncertainty belongs to every proposition (for $i> 0$).
\end{itemize}

\subsection{Neutrosophic Logic connectors}

One notes the neutrosophic logical values of the propositions $A_1$ and $A_2$ by 
$NL(A_1) = ( T_1, I_1, F_1 )$ and $NL(A_2) = ( T_2, I_2, F_2 )$.  
If, after calculations, in the below operations one obtains values $< 0$ or $> 1$, then one replaces them with $^-0$ or $1^+$  respectively.

\subsubsection{Negation}
 
$$NL(\neg A_1) = (\{1^+\}\boxminus T_1, \{1^+\}\boxminus I_1, \{1^+\}\boxminus F_1 )$$    

\subsubsection{Conjunction}                 
$$NL(A_1\wedge A_2) = ( T_1\boxdot T_2, I_1\boxdot I_2, F_1\boxdot F_2 )$$

\subsubsection{Weak or inclusive disjunction}
 
 $$NL(A_1\vee A_2) = ( T_1\boxplus T_2\boxminus (T_1\boxdot T_2),I_1\boxplus I_2\boxminus (I_1\boxdot I_2),F_1\boxplus F_2\boxminus (F_1\boxdot F_2) )$$

Many properties of the classical logic operators do not apply in neutrosophic logic.     
Neutrosophic logic operators (connectors) can be defined in many ways according to the needs of applications or of the problem solving.

\subsection{Neutrosophic Set}

Let $U$ be a universe of discourse, $M$ a set included in $U$.  An element $x$ from $U$ is noted with respect to the neutrosophic set $M$ as $x(T, I, F)$ and belongs to $M$ in the following way:  it is $t\%$ true in the set (degree of membership), $i\%$ indeterminate (unknown if it is in the set) (degree of indeterminacy), and $f\%$ false (degree of non-membership), where $t$ varies in $T$, $i$ varies in $I$, $f$ varies in $F$. This definition is analogue to NL, and similarly NS generalizes the fuzzy set (FS), especially the intuitionistic fuzzy set (IFS), intuitionistic set (IS), paraconsistent set (PS)
For example: $x(50,20,40)\in A$ means: with a belief of $50\%$ $x$ is in $A$, with a belief of $40\%$ $x$ is not in $A$, and the $20\%$ is undecidable 

\subsubsection{Neutrosophic Set Operators}
Let $A_1$ and $A_2$ be two sets over the universe $U$.  An element $x(T_1, I_1, F_1)\in A_1$ and $x(T_2, I_2, F_2)\in A_2$ [neutrosophic membership appurtenance to $A_1$ and respectively to $A_2$].   NS operators (similar to NL connectors) can also be defined in many ways.

\subsubsection{Complement} 

If $x(T1, I1, F1)\in A_1$ then $x( \{1^+\}\boxminus T_1, \{1^+\}\boxminus I_1, \{1^+\}\boxminus F_1 ) )\in \mathcal{C}(A_1)$.    

\subsubsection{Intersection}               
If $x(T_1, I_1, F_1)\in A_1$ and $x(T_2, I_2, F_2)\in A_2$ then $x( T_1\boxdot T_2, I_1\boxdot I_2, F_1\boxdot F_2 )\in A_1\cap A_2$.

\subsubsection{Union}
If $x(T_1, I_1, F_1)\in A_1$ and $x(T_2, I_2, F_2)\in A_2$ then 
$x( T_1\boxplus T_2\boxminus (T_1\boxdot T_2),I_1\boxplus I_2\boxminus (I_1\boxdot I_2),F_1\boxplus F_2\boxminus (F_1\boxdot F_2) )\in A_1\cup A_2$.

\subsubsection{Difference}
If $x(T_1, I_1, F_1)\in A_1$ and $x(T_2, I_2, F_2)\in A_2$ then $x(T_1\boxminus (T_1\boxdot T_2), I_1\boxminus (I_1\boxdot I_2), F_1\boxminus (F_1\boxdot F_2) )\in A_1\setminus A_2$.

\subsection{Differences between Neutrosophic Set and Intuitionistic Fuzzy Set}

\begin{itemize}
\item[a)] In NS there is no restriction on $T$, $I$, $F$, while in IFS the sum of components (or their superior limits) $= 1$;  thus NL can characterize the incomplete information (sum $< 1$), paraconsistent information (sum $> 1$).  
\item[b)] NS can distinguish, in philosophy, between absolute membership [NS(absolute membership)$=1^+$] and relative membership [NS(relativemembership)$=1$], while IFS cannot; absolute membership is membership in all possible worlds, relative membership is membership in at least one world.
\item[c)] In NS the components can be nonstandard, in IFS they donÕt.
\item[d)] NS, like dialetheism [some contradictions are true], can deal with paradoxes, NS(paradox element) $= (1,I,1)$, while IFS cannot.
\item[e)] NS operators can be defined with respect to $T$, $I$, $F$ while IFS operators are defined with respect to $T$ and $F$ only
\item[f)] $I$ can be split in NS in more subcomponents (for example in BelnapÕs four-valued logic \cite{Belnap_1977} indeterminacy is split into uncertainty and contradiction), but in IFS it cannot.
\end{itemize}

\subsection{N-norm}

Here each element $x$ and $y$ has three components: $x(t_1, i_1, f_1)$, $y(t_2, i_2, f_2)$.
We define :
\begin{equation*}
\begin{cases}
\max \{x, y\} = ( \max\{t_1, t_2\}, \max\{i_1, i_2\},\max\{f_1, f_2\})\\
\min \{x, y\} = ( \min\{t_1, t_2\}, \min\{i_1, i_2\},\min\{f_1, f_2\})
\end{cases}
\end{equation*}
An {\it{N-norm}} is a function $N_n: {([^-0, 1^+]\boxdot [^-0, 1^+]\boxdot [^-0, 1^+])}^2 \mapsto [^-0, 1^+]$, defined in neutrosophic set theory and neutrosophic logic to represent the {\it{intersection}} of two neutrosophic sets and the neutrosophic logical operator {\it{and}} respectively.  Extended to the fusion theory the N-norm will be a substitute for the conjunctive rule.
The N-norm satisfies the conditions:
\begin{itemize}
\item[a)] Boundary Conditions:   $N_n(0, 0) = 0, N_n(x, 1) = x$.
\item[b)] Commutativity: $N_n(x, y) = N_n(y, x)$.        
\item[c)] Monotonicity: If $ x \leq u$ and $y \leq v$, then $N_n(x, y)\leq N_n(u, v)$. 
\item[d)] Associativity: $N_n(N_n(x, y), z ) = N_n( x, N_n(y, z) )$.   
\end{itemize}
     
There are many functions which satisfy the N-norm conditions.  We present below the most known ones:
\begin{itemize}
\item The Algebraic Product N-norm: $N_{\text{n-algebraic}}(x, y) = x\boxdot y$
\item The Bounded N-norm:  $N_{\text{n-bounded}}(x, y) = \max\{0, x\boxplus y \boxminus 1\}$
\item  The Default (min) N-norm:  $N_{\text{n-min}}(x, y) = \min\{x, y\}$.
\end{itemize}

\subsection{N-conorm}

An N-conorm is a function, $N_c: {([^-0, 1^+]\boxdot [^-0, 1^+]\boxdot [^-0, 1^+])}^2 \mapsto [^-0, 1^+]$, defined in neutrosophic set theory and neutrosophic logic to represent the {\it{union}} of two neutrosophic sets and the neutrosophic logical operator {\it{or}} respectively.  Extended to the fusion theory the N-conorm will be a substitute for the disjunctive rule.
The N-conorm satisfies the conditions:
\begin{itemize}
\item[a)] Boundary Conditions:   $N_c(1, 1) = 1$, $N_c(x, 0) = x$.
\item[b)] Commutativity: $N_c(x, y) = N_c(y, x)$.        
\item[c)] Monotonicity: if  $x \leq u$ and $y \leq v$, then $N_c(x, y)\leq N_c(u, v)$. 
\item[d)] Associativity: $N_c(N_c(x, y), z ) = N_c( x, N_c(y, z) )$.  
\end{itemize}
      
There are many functions which satisfy the N-conorm conditions.  We present below the most known ones:
\begin{itemize}
\item The Algebraic Product N-conorm: $N_{\text{c-algebraic}}(x, y) = x\boxplus y\boxminus (x\boxdot y)$
\item The Bounded N-conorm: $N_{\text{c-bounded}}(x, y) = \min\{1, x\boxplus y\}$
 \item The Default (max) N-conorm: $N_{\text{c-max}}(x, y) = \max\{x, y\}$.
\end{itemize}
\noindent
Then, the {\it{N-norm Fusion rule}} and the {\it{N-conorm Fusion rule}} are defined as follows:
$$m_{Nn12}(A) = \sum_{\substack{X,Y\in \Theta \\ Y\cap Y=A}}N_n(m_1(X),m_2(Y)) \qquad\qquad m_{Nc12}(A) = \sum_{\substack{X,Y\in \Theta \\ Y\cup Y=A}}N_c(m_1(X),m_2(Y))$$

\section{Examples of N-norm and N-conorm Fusion rules}

Suppose one has the frame of discernment $\Theta=\{\theta_1,\theta_2,\theta_3\}$
and two sources $S_1$ and $S_2$ that provide respectively the following information (triple masses):
$m_1(\theta_1) = (0.6, 0.1, 0.3$), i.e. $S_1$ believes in $\theta_1$ with $60\%$, doesnÕt believe in $\theta_1$ with $30\%$, and is undecided about $\theta_1$ with $10\%$.  Similarly, one considers also
$$m_1(\theta_2)= (0.8, 0, 0.2) \qquad m_2(\theta_1) = (0.5, 0.3, 0.2) \qquad m_2(\theta_2) = (0.7, 0.2, 0.1)$$

Since one can have all kind of information (i.e. incomplete, paraconsistent, complete) the sum of an hypothesis components may be $<1$, $>1$, or $=1$.
We can normalize the hypothesis components by dividing each component by the sum of the components.

\subsection{Both Sources are right}
If we consider that both sources are right, then one uses the N-norm (letÕs take, as an example, the Algebraic Product) and one gets\footnote{where $\overset{\sim}{\equiv}$ denotes {\it{equality after normalization}}}:
\begin{align*}
m_{\text{Nn12}} (\theta_1) &= m_1(\theta_1)\boxdot m_2(\theta_1) = (0.6, 0.1, 0.3)\boxdot (0.5, 0.3, 0.2) \\
& = (0.6\cdot 0.5, 0.1\cdot 0.3, 0.3\cdot 0.2) = (0.30, 0.03, 0.06) \overset{\sim}{\equiv} (0.769231, 0.076923, 0.153846)
\end{align*}
\begin{align*}
m_{\text{Nn12}}(\theta_2) &= m_1(\theta_2)\boxdot m_2(\theta_2) = (0.8, 0, 0.2)\boxdot (0.7, 0.2, 0.1)\\
 &= (0.8\cdot 0.7, 0\cdot 0.2, 0.2\cdot 0.1) = (0.56, 0, 0.02) \overset{\sim}{\equiv} (0.965517, 0, 034483)
\end{align*} 
\begin{align*}
m_{\text{Nn12}} (\theta_1\cap \theta_2) &= [m_1(\theta_1) \boxdot m_2(\theta_2)]\boxplus [m_2(\theta_1)\boxdot m_1(\theta_2)] \\
&= [(0.6, 0.1, 0.3)\boxdot (0.7, 0.2, 0.1)]\boxplus [(0.8, 0, 0.2)\boxdot (0.5, 0.3, 0.2)]\\
& = (0.42, 0.02, 0.03) \boxplus (0.40, 0, 0.04) = (0.82, 0.02, 0.07) \overset{\sim}{\equiv} (0.901099, 0.021978, 0.076923)
\end{align*} 

If one finds out that $\theta_1\cap\theta_2 =\emptyset$, then one uses the DSm hybrid rule adjusted with the N-norm in order to transfer the conflicting mass to $m_{\text{Nn12}} (\theta_1\cup \theta_2) = (0.901099, 0.021978, 0.076923)$.

\subsection{One Source is right and another one is not, but we donÕt know which one}

We use the N-conorm (letÕs take, as an example, the Algebraic Product) and one gets:
\begin{align*}
m_{\text{Nc12}}(\theta_1) & = m_1(\theta_1)\boxplus m_2(\theta_1)\boxminus [m_1(\theta_1)\boxdot m_2(\theta_1)]  \\
& = (0.6, 0.1, 0.3)\boxplus (0.5, 0.3, 0.2)\boxminus [(0.6, 0.1, 0.3)\boxdot (0.5, 0.3, 0.2)] \\
& = ( 0.6+0.5-0.6\cdot 0.5, 0.1+0.3-0.1\cdot 0.3, 0.3+0.2-0.3\cdot 0.2 ) \\
& = (0.80, 0.37, 0.44) \overset{\sim}{\equiv} (0.496894, 0.229814, 0.273292)
\end{align*} 
\begin{align*}
m_{\text{Nc12}}(\theta_2) &= m_1(\theta_2)\boxplus m_2(\theta_2)\boxminus [m_1(\theta_2)\boxdot m_2(\theta_2)]  \\
&= (0.8, 0, 0.2)\boxplus (0.7, 0.2, 0.1)\boxminus [( 0.8, 0, 0.2)\boxdot (0.7, 0.2, 0.1)]\\
& = ( 0.8+0.7-0.8\cdot 0.7, 0+0.2-0\cdot 0.2, 0.2+0.1-0.2\cdot 0.1 ) \\
& = (0.94, 0.20, 0.28)  \overset{\sim}{\equiv} (0.661972, 0.140845, 0.197183)
\end{align*} 
\begin{align*}
m_{\text{Nc12}}(\theta_1\cap\theta_2) & = [m_1(\theta_1)\boxplus m_2(\theta_2)\boxminus  (m_1(\theta_1)\boxdot m_2(\theta_2))] \boxplus [m_1(\theta_2)\boxplus m_2(\theta_1)\boxminus (m_1(\theta_2)\boxdot m_2(\theta_1))] \\
& = 
[(0.6, 0.1, 0.3)\boxplus (0.7, 0.2, 0.1)\boxminus ( (0.6, 0.1, 0.3)\boxdot (0.7, 0.2, 0.1))] \\
& \qquad \boxplus [(0.8, 0, 0.2)\boxplus (0.5, 0.3, 0.2)\boxminus ((0.8, 0, 0.2)\boxdot (0.5, 0.3, 0.2))] \\
& = (0.88, 0.28, 0.37)\boxplus (0.90, 0.30, 0.36) \\
& = (1.78, 0.58, 0.73) \overset{\sim}{\equiv}(0.576052, 0.187702, 0.236246).
\end{align*} 

\section{Conclusion}

A general presentation of foundation of DSmT and its connection with neutrosophic logic has been proposed in this chapter. We proposed new rules of combination for uncertain, imprecise and highly conflicting sources of information. Several applications of DSmT have been proposed recently in the literature and show the efficiency of this new approach over classical rules based mainly on the Demspter's rule in the DST framework. In the last past of this chapter, we showed that
the combination of paradoxical, uncertain and imprecise sources of information can also be done using fuzzy and neutrosophic logics or sets together with DSmT and other fusion rules or theories.  The T-norms/conorm and N-norms/conorms help in redefining new fusion rules of combination or in improving the existing ones.

\end{document}